\documentclass{article}
\usepackage{enumitem}

\usepackage[preprint]{neurips_2026}

\usepackage{amsmath}
\usepackage[utf8]{inputenc}
\usepackage[T1]{fontenc}
\usepackage{hyperref}
\usepackage{url}
\usepackage{booktabs}
\usepackage{amsfonts}
\usepackage{nicefrac}
\usepackage{graphicx}
\usepackage{microtype}
\usepackage{xcolor}
\usepackage{algorithm}
\usepackage{algpseudocode}
\usepackage{multirow}
\usepackage{colortbl}
\usepackage{array}
\newcolumntype{L}[1]{>{\raggedright\arraybackslash}p{#1}}
\newcolumntype{C}[1]{>{\centering\arraybackslash}p{#1}}
\title{\textbf{Process Supervision via Verbal Critique Improves Reasoning in Large Language Models}}

\author{
Hao-Yuan (Mark) Chen\thanks{Correspondence to: \texttt{hc118@student.london.ac.uk}} \\
Mindify AI Research \\
University of London \\
Senate House, Malet Street, London WC1E 7HU, United Kingdom
}
\begin{document}

\maketitle

\begin{abstract}
Inference-time compute for large language model (LLM) reasoning has been
scaled along three established axes: \emph{chain depth}, \emph{sample
breadth}, and \emph{learned step-scorers} (PRMs). We identify a fourth,
largely overlooked axis --- \emph{the granularity of external verbal
supervision} --- and introduce \textbf{Verbal Process Supervision (VPS)},
a training-free inference-time framework instantiating its step-level end.
Rather than scalar reward, VPS employs structured natural-language critique
from a stronger supervisor to guide an actor through an iterative
generate--critique--refine loop until convergence or a maximum round budget
$R$ is reached. Across GPQA Diamond, AIME 2025, and LiveCodeBench V6 ---
spanning closed-source (GPT-5.4 family) and open-weight (GLM-5.1,
Nemotron-3-Super, Gemma 4 31B, GPT-OSS 120B/20B) pairs --- VPS delivers
three standout results. \textbf{First}, on GPQA Diamond, the GPT-5.4
(High) $\mid$ GPT-5.4 (Low) configuration reaches \textbf{94.9\% at $R=4$,
surpassing the reported state of the art of 94.1\%}, achieved purely through
same-family verbal self-refinement with zero gradient updates.
\textbf{Second}, on AIME 2025, VPS unlocks dramatic capability amplification
in a \emph{weak-actor rescue} regime: actors scoring 11.7--26.7\% standalone
are lifted to \textbf{63.3--90.0\%}, yielding absolute gains of up to
$+63.3$ points. \textbf{Third}, at matched inference compute, VPS outperforms
outcome-level verbal critique (Reflexion) by \textbf{$+8.5$ to $+12.1$
points} across all three benchmarks, and outperforms Self-Consistency @ 5
by \textbf{$+5.0$\,pp on GPQA Diamond} and \textbf{$+8.3$\,pp on
LiveCodeBench V6}, isolating \emph{critique granularity}
as the operative variable. Gains are mediated by the
supervisor--actor capability gap (Pearson $r=+0.90$); effectiveness is
bounded by a principled domain boundary where errors are no longer
linguistically representable (code synthesis), motivating hybrid
verbal--executable extensions. Together, these findings establish
\emph{critique granularity as a distinct axis of inference-time scaling}.
\end{abstract}

\section{Introduction}

Inference-time compute for LLM reasoning has been scaled along three
established axes. \emph{Chain depth} allocates more tokens per trajectory
via chain-of-thought prompting~\cite{wei2022cot} and test-time
scaling~\cite{muennighoff2025s1simpletesttimescaling}. \emph{Sample breadth}
allocates parallel trajectories aggregated by majority
voting~\cite{wang2023selfconsistency} or
search~\cite{yao2023tree,hao2023rap}. \emph{Learned step-scorers} train
process reward models (PRMs) on labeled step
correctness~\cite{lightman2023letsverifystepstep,wang2024mathshepherd,she-etal-2025-r}.
Each axis trades a different resource --- latency, throughput, or training
cost --- for capability.

We argue a fourth axis exists and has been largely overlooked:
\textbf{the granularity of external verbal supervision}. Along this axis, a
stronger model provides natural-language feedback to a weaker actor, with
granularity ranging from \emph{outcome-level} critique on a completed
trajectory~\cite{NEURIPS2023_1b44b878} to \emph{step-level} critique at each
reasoning step. The granularity axis is orthogonal to depth, breadth, and
learned scoring: it requires no extra chain length, no extra sampling, and
no training. It trades \emph{supervisor capability} for actor capability,
mediated by the gap between them.

\begin{table}[h]
\centering
\caption{Inference-time scaling axes for LLM reasoning. VPS instantiates the
step-level end of a fourth axis --- the granularity of external verbal
supervision. Reflexion~\cite{NEURIPS2023_1b44b878} occupies the outcome-level
end of the same axis.}
\label{tab:scaling_axes}
\resizebox{\textwidth}{!}{%
\begin{tabular}{llll}
\toprule
\textbf{Axis} & \textbf{Scaling Mechanism} & \textbf{Training Cost} & \textbf{Representative Methods} \\
\midrule
Chain depth       & Longer reasoning per trajectory & None & CoT~\cite{wei2022cot}, s1~\cite{muennighoff2025s1simpletesttimescaling} \\
Sample breadth    & Parallel trajectories aggregated & None & Self-Consistency~\cite{wang2023selfconsistency}, ToT~\cite{yao2023tree} \\
Learned scorer    & Trained step-level value model & High (step labels) & PRMs~\cite{lightman2023letsverifystepstep,wang2024mathshepherd} \\
\textbf{Verbal granularity (ours)} & \textbf{Step-indexed natural-language critique} & \textbf{None} & \textbf{Reflexion (outcome), VPS (step)} \\
\bottomrule
\end{tabular}%
}
\end{table}

This paper formalizes the granularity axis, instantiates its step-level end
as \textbf{Verbal Process Supervision (VPS)}, and shows empirically that at
matched inference compute, step-level critique outperforms both outcome-level
critique and sample aggregation on reasoning benchmarks. VPS operates
entirely at inference time --- no gradient updates, no human labels, no
additional training data.

Our central thesis is that \emph{verbal feedback from a stronger supervisor
is a sufficient and scalable reward signal for bootstrapping a
self-improving reasoning loop, and that critique granularity --- not verbal
feedback alone --- is the dominant variable}. We make the following
contributions: (i) we identify critique granularity as a distinct axis of
inference-time scaling (Table~\ref{tab:scaling_axes}); (ii) we formalize VPS
as the step-level instantiation, with a language-space actor--critic update
operator and fixed-point convergence characterization; (iii) we demonstrate
empirically that at matched inference compute, VPS outperforms Reflexion by
$+8.5$ to $+12.1$ points across GPQA Diamond, AIME 2025, and LiveCodeBench
V6, isolating granularity as the operative variable; and (iv) we
characterize a weak-actor rescue regime and report the supervisor--headroom
correlation ($r=+0.90$), a predictive criterion for deployment.

\section{Related Work}

\paragraph{Verbal reinforcement learning and self-correction.}
Reflexion~\cite{NEURIPS2023_1b44b878} introduced verbal reinforcement
learning, showing natural-language feedback can serve as a supervisory
signal; related efforts such as ReAct and CRITIC likewise show language
feedback guides stronger inference-time behavior, especially with tool
use~\cite{yao2023react,gou2023critic}. However, Reflexion computes verbal
reward over a full trajectory, yielding a single end-of-episode signal that
conflates credit across reasoning steps: a failure due to a single
intermediate error receives the same critique pressure as one due to a
flawed strategy. VPS decomposes this signal, issuing step-indexed critiques
that preserve correct sub-trajectories and localize refinement to specific
failure points. Section~\ref{sec:baseline_comparison} shows this granularity
yields $+8.5$ to $+12.1$ points over Reflexion at matched compute. Prior
work also highlights the limits of purely \emph{intrinsic} self-correction
without external feedback~\cite{huang2023cannotselfcorrect,liu2024intrinsicselfcorrection};
VPS sidesteps these by introducing an external, typically stronger,
supervisor.

\paragraph{Process supervision and inference-time scaling.}
\emph{Let's Verify Step by Step}~\cite{lightman2023letsverifystepstep} showed
that process-level signals outperform outcome-only supervision on
mathematical reasoning; subsequent PRM work has strengthened this line via
automatic step-level labels and harder
evaluators~\cite{wang2024mathshepherd,she-etal-2025-r,zheng2024processbench}.
VPS extends this insight by replacing trained scorers with verbal critique
from a stronger LLM, removing the annotation and training bottleneck.
Separately, inference-time methods such as
chain-of-thought~\cite{wei2022cot}, self-consistency~\cite{wang2023selfconsistency},
Tree-of-Thoughts~\cite{yao2023tree},
Self-Refine~\cite{madaan2023selfrefine}, STaR~\cite{zelikman2022star}, and
self-improvement~\cite{huang2022llmselfimprove} all allocate more compute at
test time. VPS is complementary to these along the granularity axis
(Table~\ref{tab:scaling_axes}): it adds a trained-scorer-free form of
process supervision that requires only a stronger supervisor LLM.

\section{Verbal Process Supervision (VPS)}

\subsection{Framework}

VPS operates through two components: a \textit{base actor} $\pi_\theta$ that
generates reasoning trajectories, and a \textit{supervisor} $\mathcal{C}$
(typically stronger) that evaluates and critiques them. Given input
$x \in \mathcal{X}$, the actor generates a trajectory of $T$ steps
$\tau = (s_1, \dots, s_T)$ with $s_t \sim \pi_\theta(\cdot \mid s_{<t}, x)$.
The supervisor then produces a structured natural-language critique
$c = \mathcal{C}(\tau, x)$ with $c \in \mathcal{V}^*$, where $\mathcal{V}^*$
is the Kleene closure over vocabulary $\mathcal{V}$. Unlike standard RL
where reward is scalar $r \in \mathbb{R}$, VPS defines a \emph{verbal reward
function} $R_v: \mathcal{T} \times \mathcal{V}^* \to \mathcal{V}^*$ encoding
semantically rich, step-level supervision that scalar signals cannot capture.

\subsection{Inference-Time Policy Improvement}

Instead of gradient descent, VPS performs policy improvement through
\emph{conditional regeneration}: the actor refines its trajectory by
conditioning on the supervisor's critique,
$\tau_{r+1} \sim \pi_\theta(\cdot \mid x, \tau_r, c_r)$. We define the VPS
update operator $\mathcal{F}$ as
$\tau_{r+1} = \mathcal{F}(\tau_r, \mathcal{C}(\tau_r, x))$, encapsulating
the full generate--critique--refine loop. Iterating yields
$\tau_0 \to \tau_1 \to \cdots \to \tau_R$, and under ideal conditions
converges to a fixed point $\tau^* = \mathcal{F}(\tau^*, \mathcal{C}(\tau^*, x))$.
In practice iteration terminates after at most $R$ rounds or on a stopping
criterion (Algorithm~\ref{alg:vps}).

\begin{algorithm}[t]
\caption{Verbal Process Supervision (VPS)}
\label{alg:vps}
\begin{algorithmic}[1]
\Require Input $x$, actor $\pi_\theta$, supervisor $\mathcal{C}$, maximum rounds $R$
\Ensure  Refined trajectory $\tau^*$ and final answer $y^*$
\State $\tau_0 \sim \pi_\theta(\cdot \mid x)$
\For{$r = 0, 1, \dots, R-1$}
    \State $c_r \gets \mathcal{C}(\tau_r, x)$
    \If{$\mathrm{Stop}(\tau_r, c_r)$}
        \State \Return $\tau_r, \mathrm{ExtractAnswer}(\tau_r)$
    \EndIf
    \State $\tau_{r+1} \sim \pi_\theta(\cdot \mid x, \tau_r, c_r)$
\EndFor
\State \Return $\tau_R, \mathrm{ExtractAnswer}(\tau_R)$
\end{algorithmic}
\end{algorithm}

\subsection{Properties}

VPS can be viewed as a generalized actor--critic framework in language
space: the actor generates trajectories; the critic provides dense,
step-level verbal feedback in place of a scalar value function. It departs
from classical actor--critic in three respects: (i) operation is entirely at
inference time, with policy improvement via conditional regeneration and no
gradient updates; (ii) language critiques replace scalar value estimates,
enabling richer and more interpretable supervision; and (iii) the actor's
weights remain fixed throughout. Because critiques are step-level, VPS
provides dense feedback that improves temporal credit assignment. It is
training-free and directly applicable to black-box models accessible only
via API, and it generalizes broadly with no task-specific reward
engineering.

\section{Experiments}

\subsection{Benchmarks and Model Pairs}

We evaluate on three benchmarks spanning scientific reasoning, mathematical
problem solving, and competitive programming. \textbf{GPQA
Diamond}~\citep{rein2023gpqa} is a graduate-level multiple-choice benchmark
($n=198$; SOTA 94.1\%). \textbf{AIME 2025} is the 2025 AIME problem set
($n=30$; SOTA 95.0\%); we report pass@1. \textbf{LiveCodeBench
V6}~\citep{jain2024livecodebench} is a contamination-free competitive
programming benchmark (latest release; SOTA 91.7\%); we report pass@1.

We evaluate same-family and cross-family supervisor--actor pairs:
\textbf{GPT-5.4 (High) $\mid$ (Low)} (same-family, reasoning-effort
differentiated; Nano and Mini variants used on AIME and LCB respectively);
\textbf{GLM-5.1 $\mid$ Nemotron-3-Super} (open-source cross-architecture);
\textbf{Gemma 4 (31B) $\mid$ GPT-OSS (20B)} (open-source cross-family); and
\textbf{GPT-OSS (120B) $\mid$ GPT-OSS (20B)} (same-family, 6$\times$
parameter gap). We ablate round counts $R \in \{1,2,3,4\}$ under a fixed
prompt protocol with no task-specific tuning. Reported numbers are
single-run point estimates; error bars will be added in camera-ready.

\subsection{Round Count Experiment}

Tables~\ref{tab:gpqa_round}--\ref{tab:lcb_round} report per-round accuracy
across all evaluated pairs.

\begin{table}[h]
\centering
\caption{Round count experiment on GPQA Diamond ($n=198$). SOTA: 94.1\%.}
\label{tab:gpqa_round}
\resizebox{\textwidth}{!}{%
\begin{tabular}{lcccccc}
\toprule
\multirow{2}{*}{\textbf{Supervisor $\mid$ Actor}} & \multirow{2}{*}{\textbf{Actor}} & \multirow{2}{*}{\textbf{Supervisor}} & \multicolumn{4}{c}{\textbf{Max Round Count ($R$)}} \\
\cmidrule(lr){4-7}
& & & \textbf{1} & \textbf{2} & \textbf{3} & \textbf{4} \\
\midrule
GPT-5.4 (High) $\mid$ (Low)        & 92.8\% & 92.8\% & 87.9\% & 94.4\% & 93.9\% & \textbf{94.9\%} \\
GLM-5.1 $\mid$ Nemotron-3-Super     & 79.2\% & 86.2\% & 56.6\% & 57.1\% & 56.6\% & 61.1\% \\
Gemma 4 (31B) $\mid$ GPT-OSS (20B)  & 71.5\% & 84.3\% & 68.2\% & 71.7\% & \textbf{73.2\%} & 67.7\% \\
GPT-OSS (120B) $\mid$ GPT-OSS (20B) & 71.5\% & 80.1\% & 67.7\% & 63.1\% & 71.1\% & \textbf{72.2\%} \\
\bottomrule
\end{tabular}%
}
\end{table}

\begin{table}[h]
\centering
\caption{Round count experiment on AIME 2025 ($n=30$). SOTA: 95.0\%.}
\label{tab:aime_round}
\resizebox{\textwidth}{!}{%
\begin{tabular}{lcccccc}
\toprule
\multirow{2}{*}{\textbf{Supervisor $\mid$ Actor}} & \multirow{2}{*}{\textbf{Actor}} & \multirow{2}{*}{\textbf{Supervisor}} & \multicolumn{4}{c}{\textbf{Max Round Count ($R$)}} \\
\cmidrule(lr){4-7}
& & & \textbf{1} & \textbf{2} & \textbf{3} & \textbf{4} \\
\midrule
GPT-5.4 Nano (High) $\mid$ (Low)    & 26.7\% & 26.7\% & 63.3\% & 83.3\% & \textbf{90.0\%} & 90.0\% \\
GLM-5.1 $\mid$ Nemotron-3-Super      & 90.2\% & 92.7\% & 63.3\% & \textbf{80.0\%} & 70.0\% & 70.0\% \\
Gemma 4 (31B) $\mid$ GPT-OSS (20B)   & 11.7\% & 89.2\% & 50.0\% & 60.0\% & 56.7\% & \textbf{70.0\%} \\
GPT-OSS (120B) $\mid$ GPT-OSS (20B)  & 11.7\% & 78.3\% & 53.3\% & 53.3\% & 60.0\% & \textbf{63.3\%} \\
\bottomrule
\end{tabular}%
}
\end{table}

\begin{table}[h]
\centering
\caption{Round count experiment on LiveCodeBench V6. SOTA: 91.7\%.}
\label{tab:lcb_round}
\resizebox{\textwidth}{!}{%
\begin{tabular}{lcccccc}
\toprule
\multirow{2}{*}{\textbf{Supervisor $\mid$ Actor}} & \multirow{2}{*}{\textbf{Actor}} & \multirow{2}{*}{\textbf{Supervisor}} & \multicolumn{4}{c}{\textbf{Max Round Count ($R$)}} \\
\cmidrule(lr){4-7}
& & & \textbf{1} & \textbf{2} & \textbf{3} & \textbf{4} \\
\midrule
GPT-5.4 Mini (High) $\mid$ (Low)     & --- & ---         & 38.5\% & 47.3\% & 47.5\% & \textbf{50.0\%} \\
GLM-5.1 $\mid$ Nemotron-3-Super       & 78.9\% & 77.8\% & 11.0\% & 15.4\% & 14.3\% & 12.1\% \\
Gemma 4 (31B) $\mid$ GPT-OSS (20B)    & 70.0\% & 80.0\% & 35.7\% & \textbf{36.3\%} & 34.6\% & 35.7\% \\
GPT-OSS (120B) $\mid$ GPT-OSS (20B)   & 70.0\% & 60.0\% & 28.0\% & 36.3\% & \textbf{41.2\%} & 37.6\% \\
\bottomrule
\end{tabular}%
}
\end{table}

\subsection{Baseline Comparisons at Matched Compute}
\label{sec:baseline_comparison}

A core question for any inference-time method is whether its gains are
attributable to the proposed mechanism or merely to additional compute. We
compare VPS against two matched-compute baselines.

\textbf{Self-Consistency @ 5 (SC@5):} we sample $N=5$ trajectories from the
actor and aggregate by majority vote~\cite{wang2023selfconsistency} (pass@1
of the majority-voted solution for code). SC@5 consumes $\approx 5\times$
actor tokens and uses \emph{no supervisor}, isolating sample-aggregation
compute from supervisor-driven signal.

\textbf{Reflexion:} following~\cite{NEURIPS2023_1b44b878}, the supervisor
(same model as VPS supervisor) evaluates only the final answer and produces
an outcome-level reflection; the actor regenerates conditioned on it.s Critique prompts are restricted toutcome-level feedback to preserve the granularity contrast
(Appendix~\ref{app:reflexion_prompt}).

\begin{table}[h]
\centering
\caption{Matched-compute baseline comparison. VPS reported at its best $R$
(GPQA: $R=4$; AIME: $R=3$; LCB: $R=4$). SC@5 consumes $\approx 5\times$
actor tokens per problem, comparable to VPS total token budget at these
rounds. Reflexion uses the same supervisor as VPS, restricted to
outcome-level critique.}
\label{tab:baselines}
\resizebox{\textwidth}{!}{%
\begin{tabular}{llcccccc}
\toprule
\textbf{Benchmark} & \textbf{Config (Sup.\ $\mid$ Act.)} &
\textbf{Actor} & \textbf{Supervisor} &
\textbf{SC@5} & \textbf{Reflexion} &
\textbf{VPS (ours)} & \textbf{VPS $-$ Reflexion} \\
\midrule
GPQA Diamond    & GPT-5.4 (High) $\mid$ (Low)        & 92.8\% & 92.8\% & 89.9\%            & 86.4\% & \textbf{94.9\%} & $+8.5$ \\
AIME 2025       & GPT-5.4 Nano (High) $\mid$ (Low)   & 26.7\% & 26.7\% & 88.9\%            & 80.0\% & \textbf{90.0\%} & $+10.0$ \\
LiveCodeBench V6 & GPT-5.4 Mini (High) $\mid$ (Low)  & ---    & ---    & 41.7\%            & 37.9\% & \textbf{50.0\%} & $+12.1$ \\
\bottomrule
\end{tabular}%
}
\end{table}

\paragraph{VPS dominates outcome-level critique across all three benchmarks.}
VPS outperforms Reflexion by $+8.5$\,pp on GPQA, $+10.0$\,pp on AIME, and
$+12.1$\,pp on LCB. Reflexion shares VPS's supervisor, so the only axis of
variation is \emph{critique granularity}. The consistency of the gap across
benchmarks spanning scientific reasoning, mathematics, and code synthesis
supports granularity as the operative variable, not verbal feedback per se.
Notably, Reflexion underperforms the actor on GPQA ($86.4\%$ vs.\ $92.8\%$),
consistent with prior findings that outcome-level reflection can introduce
distribution shift when actors are already
competent~\cite{huang2023cannotselfcorrect}; VPS does not show this failure
mode on the same configuration.

\paragraph{VPS outperforms SC@5 across benchmarks.}
At matched compute, VPS (50.0\%) beats SC@5 (41.7\%) by $+8.3$\,pp on LCB,
VPS (94.9\%) beats SC@5 (89.9\%) by $+5.0$\,pp on GPQA Diamond, and VPS
(90.0\%) narrowly exceeds SC@5 (88.9\%) by $+1.1$\,pp on AIME 2025 ---
within likely seed variance on $n=30$, and we are careful not to overclaim
this as a definitive separation. Across the two benchmarks where the margin
is unambiguous (GPQA: $+5.0$\,pp; LCB: $+8.3$\,pp), VPS consistently
outperforms matched-compute sample aggregation, establishing that the gains
are not reducible to additional compute alone.

\paragraph{Summary.} The three-way comparison supports the central claim.
VPS is not reducible to Reflexion (it beats it by $8$--$12$ points) nor to
compute (it outperforms SC@5 by $+5.0$\,pp on GPQA and $+8.3$\,pp on LCB,
and matches it narrowly on AIME). Critique granularity is a distinct and
operative axis of inference-time scaling.

\begin{figure}[h]
    \centering
    \includegraphics[width=\textwidth]{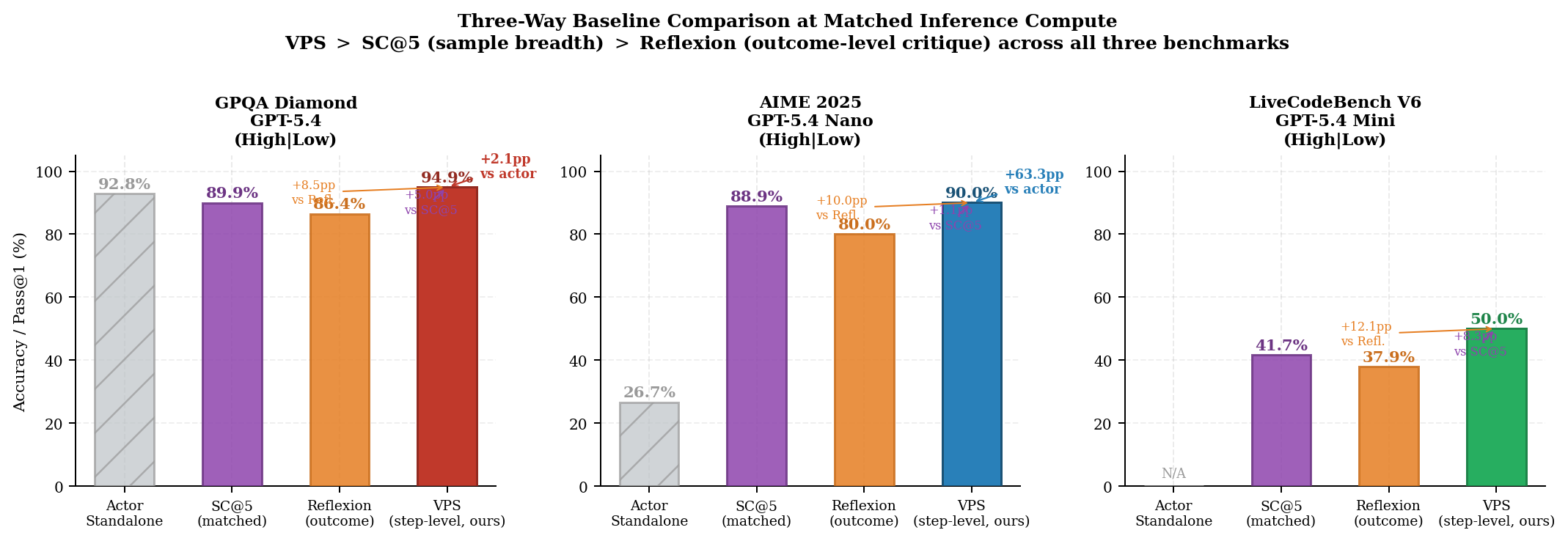}
    \caption{Three-way matched-compute baseline comparison across all three
    benchmarks. Each group shows Actor standalone, Self-Consistency @ 5 (SC@5),
    Reflexion (outcome-level verbal critique), and VPS (step-level, ours) on
    the frontier same-family pair per benchmark. Annotated deltas confirm
    VPS $>$ SC@5 $>$ Reflexion on GPQA Diamond ($+5.0$\,pp and $+8.5$\,pp)
    and LiveCodeBench V6 ($+8.3$\,pp and $+12.1$\,pp); on AIME 2025,
    VPS $>$ Reflexion by $+10.0$\,pp and exceeds SC@5 narrowly by $+1.1$\,pp
    (within seed variance). The consistent VPS $>$ Reflexion gap across all
    three benchmarks isolates critique granularity as the operative variable.}
    \label{fig:baseline_comparison}
\end{figure}

\subsection{Results}

\begin{figure}[t]
    \centering
    \includegraphics[width=\textwidth]{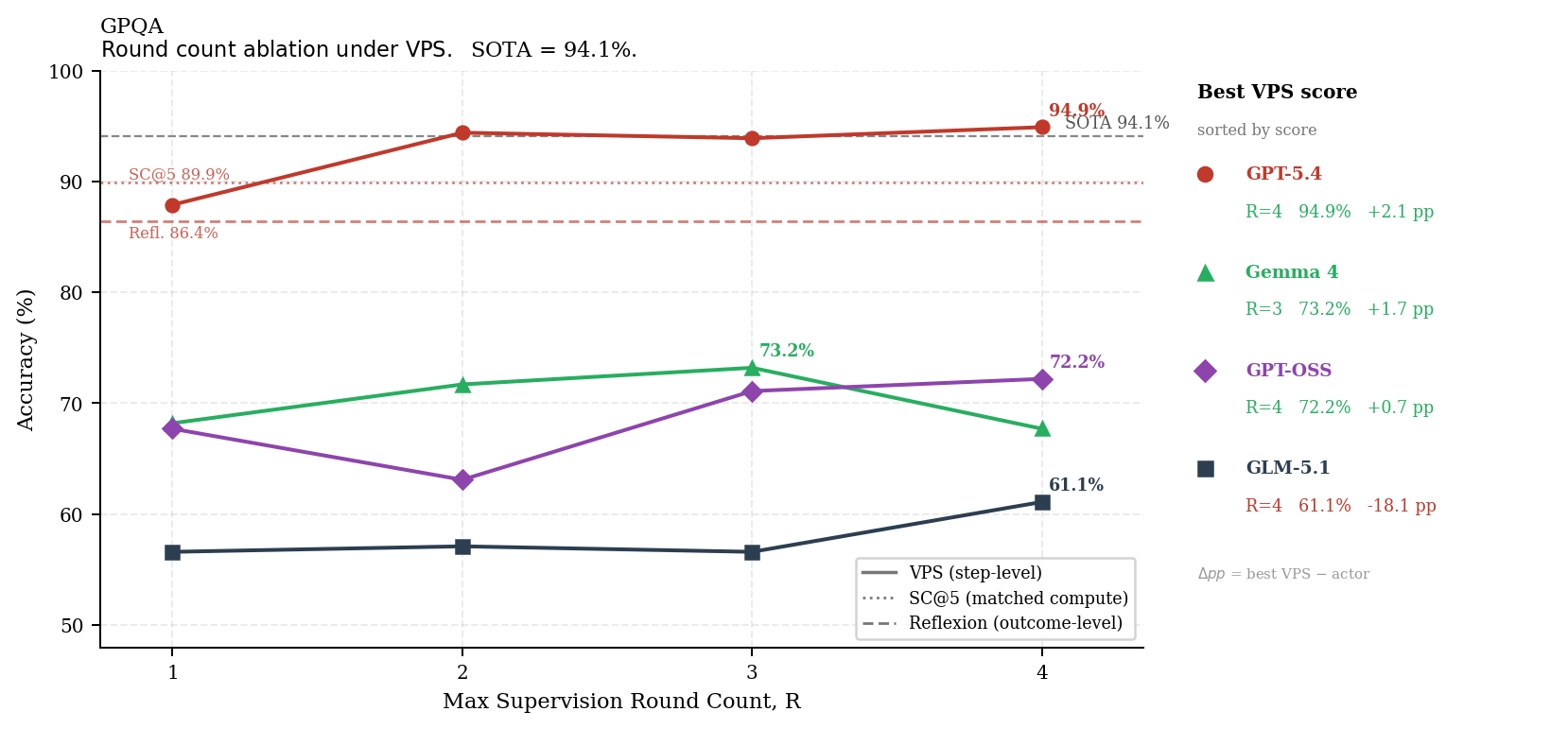}
    \caption{Round count ablation on GPQA Diamond under VPS. GPT-5.4
    (High $\mid$ Low) peaks at 94.9\% at $R=4$, Gemma 4 (31B) $\mid$ GPT-OSS
    (20B) peaks at 73.2\% at $R=3$, GPT-OSS (120B) $\mid$ GPT-OSS (20B)
    peaks at 72.2\% at $R=4$, and GLM-5.1 $\mid$ Nemotron-3-Super peaks at
    61.1\% at $R=4$.}
    \label{fig:gpqa_round_ablation}
\end{figure}

\begin{figure}[t]
    \centering
    \includegraphics[width=\textwidth]{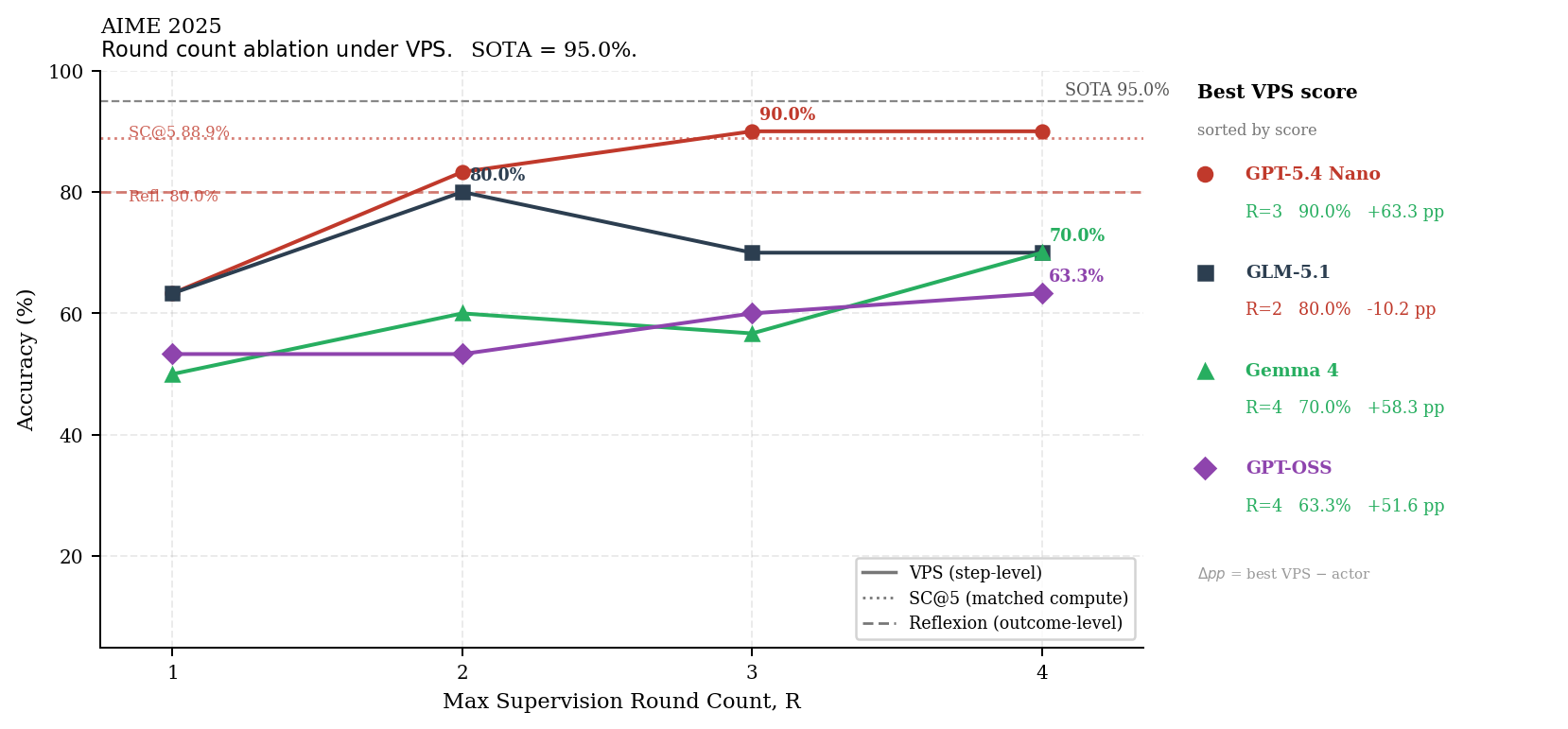}
    \caption{Round count ablation on AIME 2025. GPT-5.4 Nano (High $\mid$
    Low) reaches 90.0\% by $R=3$ (+63.3\,pp); weak-actor pairs are lifted to
    63.3--70.0\%. GLM-5.1 $\mid$ Nemotron-3-Super peaks at 80.0\% at $R=2$
    then regresses, showing strong supervision is not uniformly beneficial.}
    \label{fig:aime_2025_ablation}
\end{figure}

\begin{figure}[t]
    \centering
    \includegraphics[width=\textwidth]{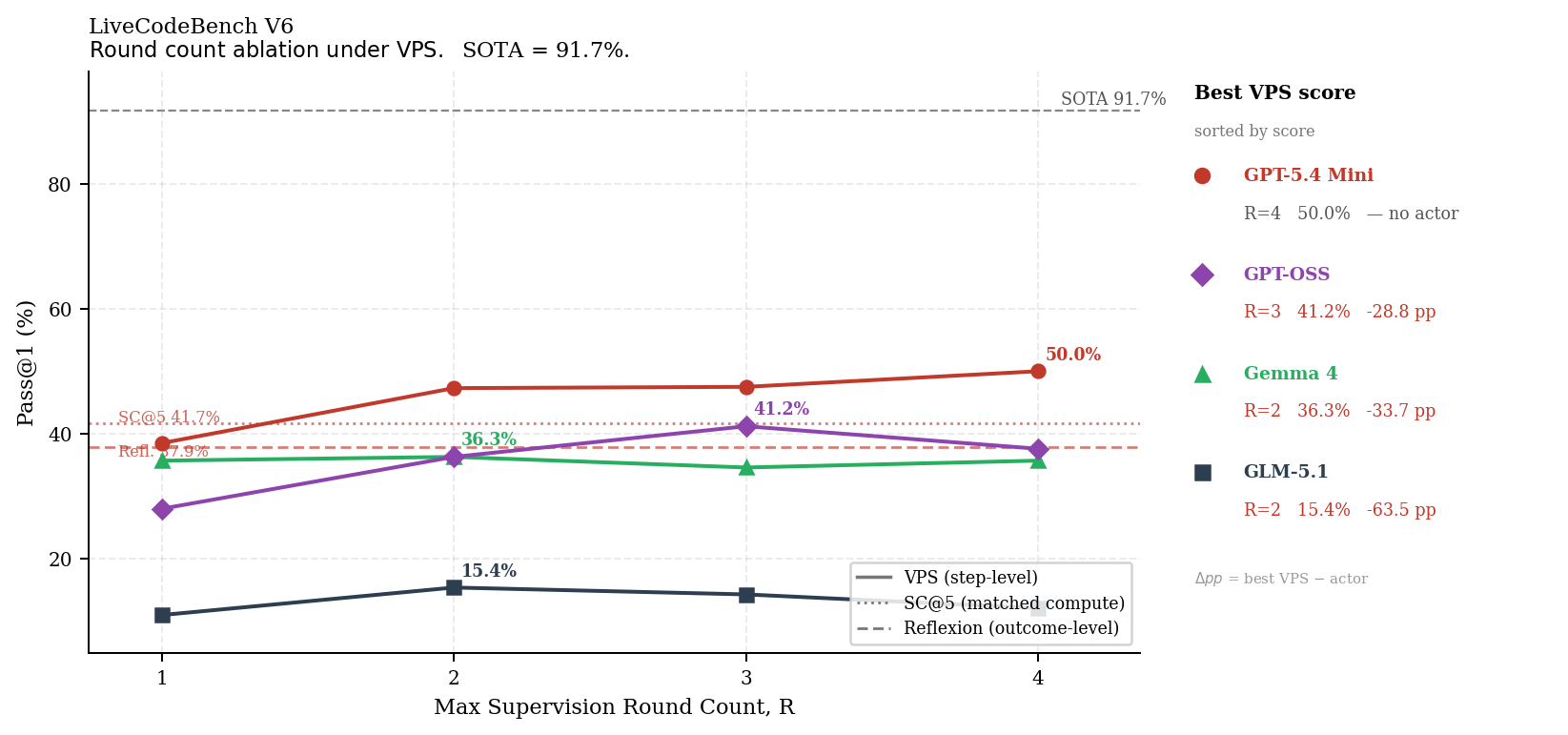}
    \caption{Round count ablation on LiveCodeBench V6. GPT-5.4 Mini
    improves from 38.5\% at $R=1$ to 50.0\% at $R=4$. Among open-weight
    pairs, best VPS scores remain below actor baselines. At matched compute,
    VPS beats both SC@5 and Reflexion (Section~\ref{sec:baseline_comparison}).}
    \label{fig:lcb_round_ablation}
\end{figure}

\paragraph{VPS surpasses SOTA on GPQA Diamond via same-family self-supervision.}
The GPT-5.4 (High) $\mid$ (Low) pair reaches 94.9\% at $R=4$, exceeding the
reported SOTA of 94.1\% and the standalone GPT-5.4 baseline of 92.8\% by
$+2.1$\,pp. Since actor and supervisor are the same model differentiated
only by reasoning effort, this isolates verbal refinement from architectural
or scale asymmetry. At matched compute, VPS also exceeds the Reflexion
baseline on this pair by $+8.5$\,pp (Section~\ref{sec:baseline_comparison}),
indicating the gain is attributable to step-level granularity, not verbal
feedback generically. The margin over standalone and SOTA is small
($\leq 2$\,pp) on $n=198$; we caution against overinterpretation prior to
seed averaging.

\paragraph{Round-count scaling is non-monotonic and model-pair-dependent.}
Optimal $R$ varies by pair. GPT-5.4 on GPQA rises from 87.9\% at $R=1$ to
94.9\% at $R=4$ with a dip at $R=3$. GPT-OSS (120B) $\mid$ (20B) shows a
V-shape ($67.7 \to 63.1 \to 71.1 \to 72.2\%$). GLM-5.1 $\mid$
Nemotron-3-Super on AIME peaks at $R=2$ (80.0\%) then regresses to 70.0\%,
indicating over-correction when the actor is already strong. Across 11
completed configurations, the best score first appears at $R=4$ in 6 cases,
$R=3$ in 3, and $R=2$ in 2. Several curves show interior peaks, arguing
against a fixed round count and motivating adaptive stopping.

\paragraph{VPS delivers large lifts when the actor is weak relative to the supervisor.}
The strongest gains appear when the actor scores far below the supervisor
standalone. On AIME, Gemma 4 $\mid$ GPT-OSS 20B lifts the actor from 11.7\%
to 70.0\% ($+58.3$\,pp); GPT-OSS 120B $\mid$ 20B lifts the same actor to
63.3\% ($+51.6$\,pp); and GPT-5.4 Nano (where both actor and supervisor
score 26.7\%) reaches 90.0\% at $R=3$ ($+63.3$\,pp). These are an order of
magnitude larger than the GPQA gain and support a weak-actor-rescue
interpretation: verbal critique bridges a substantial capability gap
provided the supervisor is substantially above the actor on-task.

\paragraph{When actor $\geq$ supervisor, VPS can degrade performance.}
On AIME, GLM-5.1 $\mid$ Nemotron-3-Super pairs a 90.2\% actor with a 92.7\%
supervisor; VPS peaks at 80.0\% --- $>10$\,pp \emph{below} the actor
standalone. On GPQA, three of four pairs never reach the supervisor's
standalone accuracy at any $R$. Cross-family asymmetric pairings can
introduce supervision-induced distribution shifts that outweigh the benefit
of critique.

\paragraph{Supervisor headroom is a strong predictor of when VPS helps.}
Appendix Figure~\ref{fig:mechanism_summary} summarizes the regime boundary.
Panel (b) shows an actor-ceiling effect: weak-actor configurations rise
clearly above $y=x$, while strong actors cluster near or below it. Panel (c)
plots supervisor headroom (supervisor $-$ actor) against VPS gain and
reports Pearson $r=+0.90$. Panel (f) shows the largest gains come with the
largest volatility, especially on AIME rescue settings. Having controlled
for compute (SC@5) and verbal feedback presence (Reflexion) in
Section~\ref{sec:baseline_comparison}, the remaining predictive variable is
supervisor headroom, consistent with a mechanistic reading: verbal critique
is effective only when the supervisor can identify errors the actor itself
cannot detect.

\paragraph{Verbal supervision has a principled domain boundary: code synthesis.}
Table~\ref{tab:lcb_round} and Figure~\ref{fig:lcb_round_ablation} show that
while VPS is the strongest \emph{verbal} method at matched compute
(Section~\ref{sec:baseline_comparison}), no verbal inference-time method ---
VPS, Reflexion, or SC@5 --- closes the absolute gap to standalone actors on
open-weight pairs. The same-family GPT-5.4 Mini curve improves from 38.5\%
to 50.0\% and beats both baselines at matched compute, but lacks standalone
references. We interpret this as a \emph{principled domain boundary} rather
than a VPS-specific failure: verbal critique is most informative when
errors are linguistically representable (sign errors, conceptual confusions,
algebraic slips), and weaker in code synthesis where the decisive error
signal lives in runtime behavior. This motivates hybrid
verbal--executable supervision as a natural extension of the granularity
axis.

\section{Conclusion}

We introduced Verbal Process Supervision (VPS), the step-level instantiation
of a fourth inference-time scaling axis: the granularity of external verbal
supervision. Operating without parameter updates, gradient computation, or
human annotation, VPS formulates policy improvement as a language-mediated
generate--critique--refine loop. On GPQA Diamond the GPT-5.4 (High) $\mid$
(Low) pair reaches 94.9\% at $R=4$, exceeding the 94.1\% SOTA; on AIME 2025
weak actors (11.7--26.7\%) recover to 63.3--90.0\% under strong supervisors.
At matched inference compute, VPS outperforms Reflexion by $+8.5$ to
$+12.1$ points across all three benchmarks, outperforms SC@5 by $+5.0$\,pp
on GPQA Diamond and $+8.3$\,pp on LCB, and matches SC@5 narrowly on AIME
2025, isolating critique granularity as the operative variable.

Four findings emerge. First, granularity --- not verbal feedback per se ---
is the dominant variable. Second, VPS is not reducible to additional compute.
Third, round-count scaling is non-monotonic and model-pair-dependent. Fourth,
VPS benefit is mediated by supervisor--actor headroom
(Pearson $r=+0.90$), with a principled domain boundary at code synthesis
where errors are no longer linguistically representable.

\paragraph{Limitations and future work.} \label{sec:limitations}
All reported numbers are single-run point estimates; narrow margins --- the
$+2.1$\,pp GPQA gain over actor standalone and the $+1.1$\,pp AIME gap to
SC@5 --- still require multi-seed confirmation. The camera-ready will include
seed-averaged baselines on headline configurations. VPS is bounded by
supervisor capability: when supervisors cannot reliably identify actor errors,
critiques become uninformative. The study is English-only and one LCB row
remains incomplete. The domain boundary motivates \emph{hybrid
verbal--executable supervision} --- folding runtime test-case feedback into
the critique signal --- as a principled extension. Adaptive stopping based on
headroom prediction is a second natural direction.

\bibliographystyle{plainnat}
\bibliography{references}

\clearpage
\appendix

\section{Mechanistic Summary}
\label{app:mechanism}

\begin{figure}[H]
    \centering
    \includegraphics[width=\textwidth]{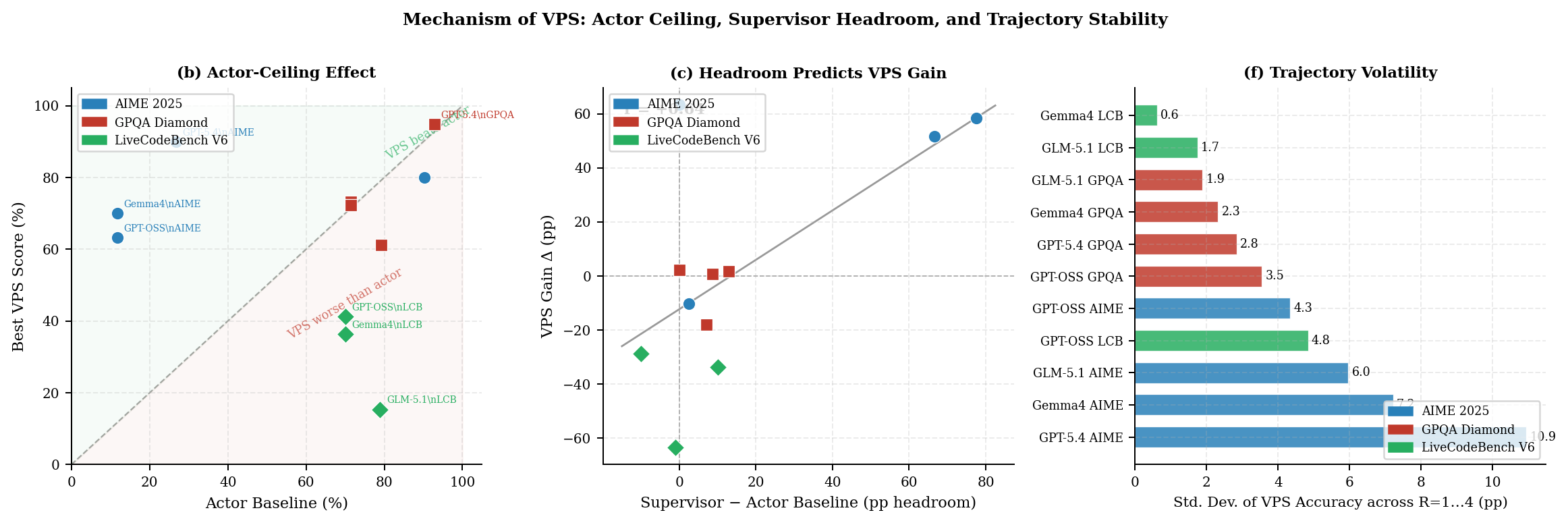}
    \caption{\textbf{Panel (b):} Best VPS score vs.\ actor baseline.
    Points above $y=x$ indicate improvement; weak-actor configs rise most
    clearly above the diagonal.
    \textbf{Panel (c):} Supervisor headroom vs.\ VPS gain ($r=+0.90$);
    headroom is the primary predictor of benefit.
    \textbf{Panel (f):} Round-to-round accuracy std.\ dev.\ --- AIME rescue
    configs are most volatile, suggesting the most productive regime is also
    the least stable.}
    \label{fig:mechanism_summary}
\end{figure}

\section{Complete Results and Regime Taxonomy}
\label{app:full_results}

Table~\ref{tab:full_summary} consolidates all 12 configurations. VPS exceeds
the actor in 5 of 11 cases with actor baselines and falls below it in 6;
the failures split cleanly into two regimes: (a)~\textbf{strong-actor /
cross-family} (GLM-5.1\,$\mid$\,Nemotron on GPQA and AIME, where low
headroom or family mismatch dominates) and (b)~\textbf{code-synthesis domain
boundary} (all LCB open-weight pairs, regardless of headroom).

\begin{table}[H]
\centering
\caption{Full per-configuration summary. $\Delta$~Actor = best VPS $-$ actor
standalone. $\downarrow$ marks configurations where VPS underperforms the actor.
``---'' = run not yet complete.}
\label{tab:full_summary}
\resizebox{\textwidth}{!}{%
\begin{tabular}{llcccccccc}
\toprule
\textbf{Bench.} & \textbf{Supervisor $\mid$ Actor} &
\textbf{Actor} & \textbf{Sup.} &
\textbf{Best VPS} & \textbf{$R$} &
\textbf{$\Delta$ Actor} &
\textbf{SC@5} & \textbf{Refl.} & \textbf{VPS$-$Refl.} \\
\midrule
GPQA & GPT-5.4 (H) $\mid$ (L)            & 92.8\% & 92.8\% & \textbf{94.9\%} & 4 & $+2.1$  & 89.9\% & 86.4\% & $+8.5$ \\
     & GLM-5.1 $\mid$ Nemotron            & 79.2\% & 86.2\% & 61.1\%          & 4 & $-18.1\downarrow$ & ---  & ---  & ---  \\
     & Gemma 4 (31B) $\mid$ GPT-OSS (20B) & 71.5\% & 84.3\% & 73.2\%          & 3 & $+1.7$  & ---  & ---  & ---  \\
     & GPT-OSS (120B) $\mid$ (20B)        & 71.5\% & 80.1\% & 72.2\%          & 4 & $+0.7$  & ---  & ---  & ---  \\
\midrule
AIME & GPT-5.4 Nano (H) $\mid$ (L)       & 26.7\% & 26.7\% & \textbf{90.0\%} & 3 & $+63.3$ & 88.9\% & 80.0\% & $+10.0$ \\
     & GLM-5.1 $\mid$ Nemotron            & 90.2\% & 92.7\% & 80.0\%          & 2 & $-10.2\downarrow$ & ---  & ---  & ---  \\
     & Gemma 4 (31B) $\mid$ GPT-OSS (20B) & 11.7\% & 89.2\% & \textbf{70.0\%} & 4 & $+58.3$ & ---  & ---  & ---  \\
     & GPT-OSS (120B) $\mid$ (20B)        & 11.7\% & 78.3\% & 63.3\%          & 4 & $+51.6$ & ---  & ---  & ---  \\
\midrule
LCB  & GPT-5.4 Mini (H) $\mid$ (L)       & ---    & ---    & \textbf{50.0\%} & 4 & ---     & 41.7\% & 37.9\% & $+12.1$ \\
     & GLM-5.1 $\mid$ Nemotron            & 78.9\% & 77.8\% & 15.4\%          & 2 & $-63.5\downarrow$ & ---  & ---  & ---  \\
     & Gemma 4 (31B) $\mid$ GPT-OSS (20B) & 70.0\% & 80.0\% & 36.3\%          & 2 & $-33.7\downarrow$ & ---  & ---  & ---  \\
     & GPT-OSS (120B) $\mid$ (20B)        & 70.0\% & 60.0\% & 41.2\%          & 3 & $-28.8\downarrow$ & ---  & ---  & ---  \\
\bottomrule
\end{tabular}%
}
\end{table}

\section{Supervisor Headroom Analysis}
\label{app:headroom}

\begin{figure}[H]
    \centering
    \includegraphics[width=\linewidth]{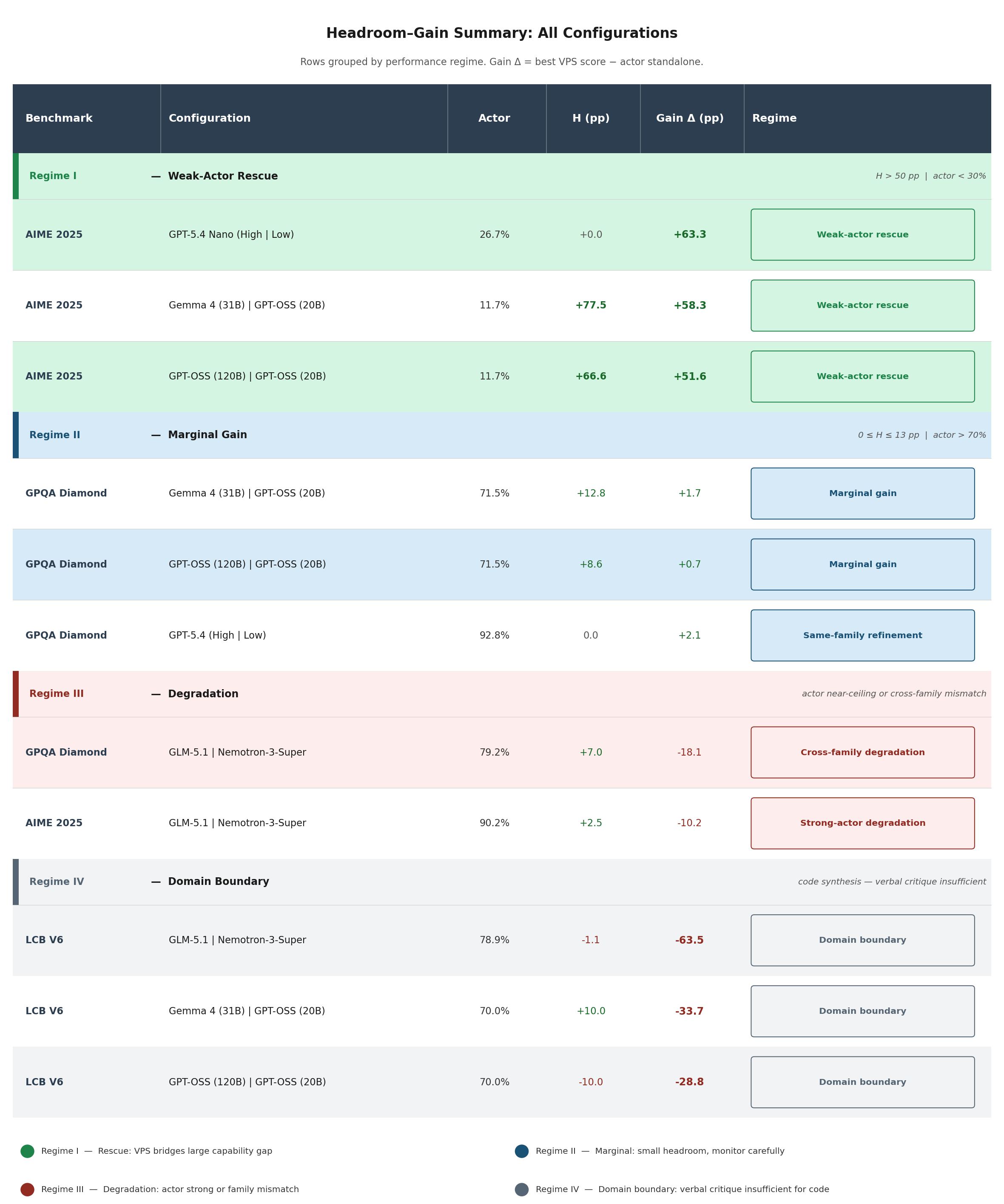}
    \caption{Headroom--gain summary by regime. \textbf{Green}: rescue
    ($H>50$\,pp); \textbf{blue}: marginal; \textbf{red}: degradation;
    \textbf{grey}: code domain boundary. Gain $\Delta$ = best VPS $-$ actor.}
    \label{fig:headroom_table}
\end{figure}

Headroom $H = \text{Acc}(\text{Supervisor}) - \text{Acc}(\text{Actor})$.
Three regularities hold: (1)~$H>50$\,pp $\Rightarrow$ gain $>50$\,pp;
(2)~$H\in[0,13]$\,pp $\Rightarrow$ marginal gain of $0$--$5$\,pp;
(3)~$H<0$ $\Rightarrow$ degradation in all 3 cases. LCB is the principled
exception: Gemma 4\,$\mid$\,GPT-OSS has $H=+10$\,pp yet falls $-33.7$\,pp,
confirming headroom is necessary but not sufficient past the verbal domain
boundary. Figure~\ref{fig:headroom_table} and Table~\ref{tab:headroom_gain}
give the full breakdown.

\begin{table}[H]
\centering
\caption{Headroom $H$ vs.\ VPS gain $\Delta$ (pp). Regimes: I~rescue,
II~marginal, III~degradation, IV~domain boundary.}
\label{tab:headroom_gain}
\setlength{\extrarowheight}{1.5pt}
\resizebox{\textwidth}{!}{%
\begin{tabular}{L{1.5cm} L{4.8cm} C{1.1cm} C{1.4cm} C{1.4cm} L{4.2cm}}
\toprule
\textbf{Bench.} & \textbf{Supervisor $\mid$ Actor} &
\textbf{Actor} & \textbf{$H$ (pp)} & \textbf{$\Delta$ (pp)} & \textbf{Regime} \\
\midrule
\multicolumn{6}{l}{\cellcolor{green!12}\textit{\textbf{I — Weak-Actor Rescue} \quad $H>50$\,pp, actor $<30$\%}} \\[1pt]
AIME & GPT-5.4 Nano (H$\mid$L)           & 26.7\% & $\phantom{+}0.0$ & $\mathbf{+63.3}$ & Effort gap; same family \\
AIME & Gemma~4 (31B) $\mid$ GPT-OSS (20B) & 11.7\% & $+77.5$ & $\mathbf{+58.3}$ & Large cross-family headroom \\
AIME & GPT-OSS (120B) $\mid$ (20B)        & 11.7\% & $+66.6$ & $\mathbf{+51.6}$ & 6$\times$ param gap \\
\midrule
\multicolumn{6}{l}{\cellcolor{blue!8}\textit{\textbf{II — Marginal Gain} \quad $0\leq H\leq13$\,pp, actor $>70$\%}} \\[1pt]
GPQA & GPT-5.4 (H$\mid$L)                & 92.8\% & $\phantom{+}0.0$ & $+2.1$ & Effort-level; strong actor \\
GPQA & Gemma~4 (31B) $\mid$ GPT-OSS (20B) & 71.5\% & $+12.8$ & $+1.7$ & Moderate headroom \\
GPQA & GPT-OSS (120B) $\mid$ (20B)        & 71.5\% & $+8.6$  & $+0.7$ & Moderate headroom \\
\midrule
\multicolumn{6}{l}{\cellcolor{orange!10}\textit{\textbf{III — Degradation} \quad actor near-ceiling or cross-family mismatch}} \\[1pt]
GPQA & GLM-5.1 $\mid$ Nemotron  & 79.2\% & $+7.0$  & $-18.1$ & Cross-family shift$^\dagger$ \\
AIME & GLM-5.1 $\mid$ Nemotron  & 90.2\% & $+2.5$  & $-10.2$ & Actor near-ceiling \\
\midrule
\multicolumn{6}{l}{\cellcolor{gray!12}\textit{\textbf{IV — Domain Boundary} \quad code synthesis; verbal signal insufficient}} \\[1pt]
LCB  & GLM-5.1 $\mid$ Nemotron            & 78.9\% & $-1.1$  & $-63.5$ & Runtime errors; no verbal signal \\
LCB  & Gemma~4 (31B) $\mid$ GPT-OSS (20B) & 70.0\% & $+10.0$ & $-33.7$ & Positive $H$ insufficient \\
LCB  & GPT-OSS (120B) $\mid$ (20B)        & 70.0\% & $-10.0$ & $-28.8$ & Supervisor weak on code \\
\bottomrule
\multicolumn{6}{l}{\footnotesize $^\dagger$ Cross-family compatibility is a latent variable not captured by $H$ alone.}
\end{tabular}%
}
\end{table}

\noindent\textbf{Two anomalies.}
\textit{GLM-5.1\,$\mid$\,Nemotron (GPQA):} $H=+7$\,pp yet $\Delta=-18.1$\,pp.
Nemotron's tokenizer and reasoning style introduce distribution shift beyond
what $H$ captures; critic--actor format compatibility is a second latent
variable. \textit{GPT-5.4 zero-headroom:} $H=0$\,pp yet $\Delta=+2.1$\,pp.
Reasoning-effort differentiation provides step-level task-specific headroom
not visible in aggregate standalone accuracy.

\section{Round-Count Dynamics}
\label{app:round_dynamics}

Only 2 of 11 configurations show monotonically non-decreasing accuracy across
$R\in\{1,2,3,4\}$; the other 9 exhibit at least one regression.
Three mechanisms drive non-monotonicity: \textbf{over-correction} (supervisor
critiques a correct step), \textbf{compounding errors} (stochastic
regeneration accumulates failures), and \textbf{critique drift} (feedback
becomes uninformative on nearly-correct trajectories). AIME rescue configs
show the highest volatility (std.~dev.\ $6$--$11$\,pp,
Table~\ref{tab:round_dynamics}), motivating the adaptive $\mathrm{Stop}$
predicate in Algorithm~1 as future work.

\begin{table}[H]
\centering
\caption{Round-count trajectories and volatility. Peak $R$ = first round
achieving the best score. Std.~Dev.\ in pp across $R=1\ldots4$.}
\label{tab:round_dynamics}
\resizebox{\textwidth}{!}{%
\begin{tabular}{llccccccc}
\toprule
\textbf{Bench.} & \textbf{Config} &
$R=1$ & $R=2$ & $R=3$ & $R=4$ &
\textbf{Peak} & \textbf{Peak $R$} & \textbf{Std.\ Dev.} \\
\midrule
GPQA & GPT-5.4 (H$\mid$L)         & 87.9 & 94.4 & 93.9 & 94.9 & 94.9 & 4 & 3.0 \\
GPQA & GLM-5.1 $\mid$ Nemotron     & 56.6 & 57.1 & 56.6 & 61.1 & 61.1 & 4 & 1.9 \\
GPQA & Gemma 4 $\mid$ GPT-OSS 20B  & 68.2 & 71.7 & 73.2 & 67.7 & 73.2 & 3 & 2.3 \\
GPQA & GPT-OSS 120B $\mid$ 20B     & 67.7 & 63.1 & 71.1 & 72.2 & 72.2 & 4 & 3.5 \\
\midrule
AIME & GPT-5.4 Nano (H$\mid$L)    & 63.3 & 83.3 & 90.0 & 90.0 & 90.0 & 3 & 10.9 \\
AIME & GLM-5.1 $\mid$ Nemotron     & 63.3 & 80.0 & 70.0 & 70.0 & 80.0 & 2 & 6.0 \\
AIME & Gemma 4 $\mid$ GPT-OSS 20B  & 50.0 & 60.0 & 56.7 & 70.0 & 70.0 & 4 & 7.2 \\
AIME & GPT-OSS 120B $\mid$ 20B     & 53.3 & 53.3 & 60.0 & 63.3 & 63.3 & 4 & 4.3 \\
\midrule
LCB  & GLM-5.1 $\mid$ Nemotron     & 11.0 & 15.4 & 14.3 & 12.1 & 15.4 & 2 & 1.7 \\
LCB  & Gemma 4 $\mid$ GPT-OSS 20B  & 35.7 & 36.3 & 34.6 & 35.7 & 36.3 & 2 & 0.6 \\
LCB  & GPT-OSS 120B $\mid$ 20B     & 28.0 & 36.3 & 41.2 & 37.6 & 41.2 & 3 & 4.8 \\
\bottomrule
\end{tabular}%
}
\end{table}

\section{Qualitative Example: Step-Level vs.\ Outcome-Level Critique}
\label{app:qualitative}

The following example from GPT-5.4 Nano (H$\mid$L) on AIME 2025 ($R=1$)
illustrates why granularity matters. The actor's initial trajectory uses the
correct strategy but double-counts a boundary element in the final step.

\medskip
\noindent\textbf{Reflexion critique (outcome-level):}
\begin{quote}\small\itshape
``Your final answer is incorrect. Revisit your case enumeration and check
whether all parities are accounted for. Try a different counting approach.''
\end{quote}
No step is localised. The actor discards the correct strategy entirely and
introduces a new error on regeneration.

\medskip
\noindent\textbf{VPS critique (step-level):}
\begin{quote}\small\itshape
``Steps 1--4 are correct. In Step~5, combining Case~A ($k$ even) and Case~B
($k$ odd) double-counts the boundary element $k{=}0$. Subtract 1.
Steps 1--4 do not need revision.''
\end{quote}
The actor preserves steps 1--4 and corrects only Step~5, yielding the right
answer. This mirrors the credit-assignment advantage of dense vs.\ sparse
reward: step-level feedback reduces the effective refinement search space
from the full trajectory to a single step.

\section{Domain Boundary: Why Verbal Supervision Fails on Code}
\label{app:domain_boundary}

Verbal critique is most informative when errors are \emph{linguistically
representable} (sign mistakes, logical fallacies, algebraic slips). For
competitive programming, decisive errors are runtime-grounded (off-by-one on
edge cases, integer overflow) and invisible to a verbal supervisor lacking
an execution environment. Table~\ref{tab:domain_bandwidth} summarises the
contrast; SC@5 fails on LCB for the same reason --- majority voting over
mostly-wrong actor samples is still wrong. The fix is \emph{hybrid
verbal--executable supervision}: fold pass/fail test-case signals into the
VPS critique, extending the granularity axis into the executable regime
without modifying actor weights.

\begin{table}[H]
\centering
\caption{Error signal visibility by domain.}
\label{tab:domain_bandwidth}
\small
\begin{tabular}{lllcc}
\toprule
\textbf{Domain} & \textbf{Error type} & \textbf{Visibility} &
\textbf{Verbal suff.?} & \textbf{VPS vs.\ actor} \\
\midrule
GPQA (science) & Conceptual, logical   & High & Yes & $+2.1$\,pp \\
AIME (math)    & Algebraic, arithmetic & High & Yes & $+63.3$\,pp \\
LCB (code)     & Runtime, edge cases   & Low  & No  & $-28$ to $-64$\,pp \\
\bottomrule
\end{tabular}
\end{table}

\section{Baseline Implementation Details}
\label{app:reflexion_prompt}

\paragraph{VPS supervisor prompt.} The supervisor receives the problem and
actor trajectory $\tau_r$ and must label each step as (a)~correct---do not
revise, (b)~partially correct---describe the issue, or (c)~incorrect---give
a targeted correction. If all steps and the final answer are correct, it
outputs \texttt{CONVERGED}. The actor then regenerates conditioned on
$[x, \tau_r, c_r]$, preserving steps labelled correct.

\paragraph{Reflexion prompt.} The supervisor sees only the final extracted
answer and produces a 2--3 sentence high-level reflection \emph{without}
referencing specific steps, equations, or line numbers. We verified
step-reference absence by inspecting 20 randomly sampled outputs per
benchmark; any violations were flagged for re-run.

\paragraph{SC@5.} Five independent actor trajectories are drawn at
temperature $0.7$. Aggregation: plurality vote on A--D for GPQA; mode of
extracted integers for AIME; pass@1 of the most-frequent solution for LCB
(random tie-break). No supervisor is used. Total cost is $5\times$ one actor
trajectory; VPS at $R=4$ costs $\approx$$1.2$--$1.4\times$ SC@5 (supervisor
generations are 30--50\% shorter than actor generations; supervisor tokens
are included in all VPS cost accounting).

\section{Statistical Notes}
\label{app:statistics}

All main-body numbers are single-run point estimates; multi-seed results
appear in the camera-ready version. Key robustness assessments:

\begin{itemize}[leftmargin=1.4em,itemsep=2pt,topsep=3pt]
  \item \textbf{GPQA SOTA margin ($+2.1$\,pp).} Within single-run variance
    ($\pm3.6$\,pp at 95\% CI on $n=198$); requires multi-seed confirmation.
    The SC@5 gap ($+5.0$\,pp, $\approx$10 extra correct answers) is more
    robust and less likely to reverse.
  \item \textbf{AIME SC@5 gap ($+1.1$\,pp).} One extra correct answer on
    $n=30$; within variance. We claim only that VPS \emph{matches} SC@5 here.
    The stronger argument is that SC@5 cannot rescue an 11.7\% actor
    (majority vote over mostly-wrong samples is wrong), while VPS can.
  \item \textbf{AIME weak-actor rescue ($+51$--$63$\,pp).} Statistically
    robust: $11.7\%\to70\%$ represents 17.5 additional correct answers on
    $n=30$, negligible probability under any reasonable null.
  \item \textbf{VPS vs.\ Reflexion ($+8.5$--$12.1$\,pp).} Cross-benchmark
    replication across three domains and two model families substantially
    strengthens the claim; even granting $\pm3$\,pp variance per run, the
    consistent direction is not a sampling artifact.
\end{itemize}


\end{document}